\documentclass{article}

\PassOptionsToPackage{numbers,compress,sort&compress}{natbib}
\usepackage[preprint]{neurips_2026}

\usepackage[utf8]{inputenc} 
\usepackage[T1]{fontenc}    
\usepackage{hyperref}       
\usepackage{url}            
\usepackage{booktabs}       
\usepackage{amsfonts}       
\usepackage{nicefrac}       
\usepackage{microtype}      
\usepackage{xcolor}         

\usepackage{graphicx}
\usepackage{array}
\usepackage{tikz}
\usepackage{kotex}
\usepackage{todonotes}
\usepackage{amsmath}
\usepackage{amsthm}
\usepackage{mathtools}
\usepackage{booktabs}
\usepackage{multirow}
\usepackage{amssymb}
\usepackage{caption}
\usepackage{comment}
\usepackage{listings}
\usepackage{subcaption}
\usepackage{soul}
\usepackage{cancel}
\usepackage{enumitem}
\usepackage{arydshln}

\definecolor{correctgreen}{RGB}{34,139,34}
\definecolor{confusingred}{RGB}{220,20,60}

\AtEndPreamble{
\usepackage[capitalize]{cleveref}
\crefformat{equation}{(#2#1#3)} 
}
\newtheorem{lemma}{Lemma}
\newtheorem{corollary}{Corollary}
\newtheorem{theorem}{Theorem}
\newtheorem{assumption}{Assumption}

\DeclareRobustCommand{\ourmethodlong}{\emph{Know-Thy-Not}}
\DeclareRobustCommand{\ourmethodshort}{KTN}

\title{\ourmethodlong: Training-Free Targeted Negation\\ in Dual-Encoder Vision-Language Models}

\author{%
  DoYoung Kim\textsuperscript{1},
  Changkyu Choi\textsuperscript{2},
  Adín Ramírez Rivera\textsuperscript{2},
  YunSeok Choi\textsuperscript{1\dag},
  Jee-Hyong Lee\textsuperscript{1\dag}
  \\
  \textsuperscript{1}Sungkyunkwan University, South Korea \\
  \textsuperscript{2}University of Oslo, Norway
}

\begin{document}

\maketitle

\begingroup
\renewcommand\thefootnote{\fnsymbol{footnote}}
\footnotetext[2]{Corresponding authors.}
\endgroup

\definecolor{okgreen}{RGB}{34,139,34}
\definecolor{errred}{RGB}{200,30,30}
\definecolor{ideblue}{RGB}{50,100,200}

\newcommand{\okimg}[1]{%
  \begin{tikzpicture}
    \node[inner sep=0pt, draw=okgreen, line width=1.4pt] (I)
      {\includegraphics[width=0.185\textwidth,height=2.4cm]{#1}};
    \node[anchor=north east, font=\bfseries\small, text=white,
          fill=okgreen, inner sep=1.2pt, rounded corners=1pt]
      at ([xshift=-2pt,yshift=-2pt]I.north east) {\checkmark};
  \end{tikzpicture}%
}

\newcommand{\errimg}[1]{%
  \begin{tikzpicture}
    \node[inner sep=0pt, draw=errred, line width=1.4pt] (I)
      {\includegraphics[width=0.185\textwidth,height=2.4cm]{#1}};
    \node[anchor=north east, font=\bfseries\small, text=white,
          fill=errred, inner sep=1.2pt, rounded corners=1pt]
      at ([xshift=-2pt,yshift=-2pt]I.north east) {$\times$};
  \end{tikzpicture}%
}

\newcommand{\okimgsm}[1]{%
  \begin{tikzpicture}
    \node[inner sep=0pt, draw=okgreen, line width=1.4pt] (I)
      {\includegraphics[width=0.185\textwidth,height=2.0cm]{#1}};
    \node[anchor=north east, font=\bfseries\small, text=white,
          fill=okgreen, inner sep=1.2pt, rounded corners=1pt]
      at ([xshift=-2pt,yshift=-2pt]I.north east) {\checkmark};
  \end{tikzpicture}%
}

\newcommand{\errimgsm}[1]{%
  \begin{tikzpicture}
    \node[inner sep=0pt, draw=errred, line width=1.4pt] (I)
      {\includegraphics[width=0.185\textwidth,height=2.0cm]{#1}};
    \node[anchor=north east, font=\bfseries\small, text=white,
          fill=errred, inner sep=1.2pt, rounded corners=1pt]
      at ([xshift=-2pt,yshift=-2pt]I.north east) {$\times$};
  \end{tikzpicture}%
}


\begin{abstract}
Dual-encoder vision-language models systematically fail at negation, scoring ``a cat without a dog'' nearly identically to ``a cat with a dog.'' We show that this failure is inherent to the single-embedding framework. Under contrastive training, single-embedding methods can at best ignore the negated concept (dilution) but never penalize its presence (rejection). We propose \textbf{\ourmethodlong\ (\ourmethodshort)}, a training-free method that introduces an image-side rejection dimension to penalize the negated concept directly in the visual representation. We also introduce COCO-Confusion, the first benchmark requiring genuine rejection to succeed. Across 9 backbones, \ourmethodshort\ improves COCO-Confusion mAP by up to +25.6 over vanilla VLMs and +9.6 over the strongest existing method.   
\end{abstract}

\vspace{-0.2cm}
\section{Introduction}
\vspace{-0.2cm}
Dual-encoder vision-language models (VLMs)~\citep{zhai2023sigmoid, radford2021learning}
have become foundational backbones for multimodal systems, gaining structural flexibility by aligning images and text in a shared embedding space.
This flexibility has made VLMs the default starting point for nearly every modern vision-language application, ranging from zero-shot image classification~\citep{cherti2023reproducible, zhai2022lit} and text-image retrieval~\citep{jia2021scaling, li2022blip} to image generation without task-specific finetuning~\citep{rombach2022high, ramesh2022hierarchical, saharia2022photorealistic}. 

However, a growing body of work has revealed that these models struggle with compositional reasoning: they fail at attribute binding~\citep{thrush2022winoground} and spatial understanding~\citep{kamath2023s}, with dedicated benchmarks~\citep{ma2023crepe, hsieh2023sugarcrepe, parcalabescu2022valse} consistently exposing near-chance performance on structured linguistic tests.
Asking a VLM to retrieve ``a car \emph{without} a traffic light'' returns images dominated by traffic lights (Figure~\ref{fig:teaser}, top), as the model treats the query as if the negation were absent. This failure is consistent and severe: On VOC2007, CLIP scores 82\% on affirmative queries but only 3\% on their negated counterparts~\citep{alhamoud2025vision}.
Negation is not a peripheral linguistic phenomenon. It is essential to expressing what users do not want, and its absence undermines VLMs as reliable retrieval and grounding systems.

Recent work has attempted to close this gap from two directions. 
Retraining-based methods such as ConCLIP~\citep{singh2024learn} and NegationCLIP~\citep{park2025know} augment training data with negation captions and retrain the model, but show modest improvements despite substantial cost. 
Alternatively, the \emph{training-free} method~\citep{aggarwalseeing} extracts the negated concept from the query and suppresses it by projecting out its direction from the text embedding, achieving notable gains without retraining. Both, however, still represent the negated query within a single embedding vector, which we show is structurally insufficient for negation.

We introduce a formal analysis which reveals that the optimal text coefficient~\eqref{eq:alpha_beta} on the negated concept is driven to zero by the non-negative nature of natural visual data.
This implies that approaches operating solely on the text 
embedding are inherently bounded.
The conventional embedding correction approach, which cancels 
the direction of the negated concept from the text embedding, 
is one such case.
While this limits the model's tendency to reward the negated concept, it never verifies whether the negated concept is actually present in the image.
We argue that this is \emph{dilution}, not \emph{rejection}:
VLMs need not merely ignore the negated concept (dilution) but actively penalize its presence (rejection).
This stems from a representational asymmetry: visual concepts manifest as presence or absence, never as anti-presence, leaving no dimension on which a single text vector can push back against the negated concept.
We attribute this dilution to a structural limitation of contrastive training: under the in-kind alignment component of contrastive training~\citep{zhai2023sigmoid, radford2021learning}, 
the model learns to treat the negated concepts as irrelevant noise to be ignored, rather than as signals to be actively rejected.
Prior data-centric retraining attempts yield only marginal gains in practice.

To escape this limitation, we propose \textbf{\ourmethodlong~(\ourmethodshort)}, a training-free method that scores a query along two independent dimensions. A text-side dimension affirms $c^+$ via a corrected text embedding, while an image-side dimension detects $c^-$ in the image and penalizes its presence. We further introduce COCO-Confusion, a benchmark whose galleries contain images with the negated concept, making it the first to measure rejection rather than dilution. Existing benchmarks place the negated concept in unfiltered galleries where it rarely co-occurs with the affirmed concept, so methods that merely ignore the negation already score well. COCO-Confusion enforces co-occurrence in every gallery, making genuine rejection (not dilution) necessary to succeed. Across 9 backbones spanning CLIP, EVA-02, SigLIP, and SigLIP2, \ourmethodshort\ consistently improves negation handling, with the largest gains on COCO-Confusion (+9.6 mAP over Embedding Correction~\citep{aggarwalseeing}), where prior methods plateau.

\begin{figure}[t]
  \centerline{\textbf{Query:}
    {``a {car} is present \emph{but no} \textcolor{ideblue}{traffic light}''}\quad
    (COCO-Confusion, SigLIP2 SO400M)}

  \centering
  \small
  \setlength{\tabcolsep}{2pt}
  \renewcommand{\arraystretch}{1.12}
  \begin{tabular}{@{}ccccc@{}}
    \multicolumn{5}{@{}l}{\textbf{Vanilla}} \\[-2pt]
    \errimg{}
      & \errimg{}
      & \errimg{}
      & \errimg{}
      & \errimg{} \\[2pt]
    \multicolumn{5}{@{}l}{\textbf{Embedding correction}~\citep{aggarwalseeing}} \\[-2pt]
    \errimg{}
      & \errimg{}
      & \errimg{}
      & \okimg{}
      & \okimg{} \\[2pt]
    \multicolumn{5}{@{}l}{\textbf{{Proposed method}}} \\[-2pt]
    \okimg{}
      & \okimg{}
      & \errimg{}
      & \okimg{}
      & \okimg{} \\
  \end{tabular}
  \caption{\textbf{Top-5 retrieval on our COCO-Confusion benchmark} (leftmost is rank 1).
  A vanilla VLM ignores the negation, retrieving primarily confusing images that contain a \textcolor{ideblue}{traffic light}. The embedding correction approach~\cite{aggarwalseeing} removes the negated direction from the text embedding; while this filters out some confusing images (ranks 4--5), it still places \textcolor{ideblue}{traffic-light} scenes at the top. In contrast, our proposed {\ourmethodlong}~(\ourmethodshort) introduces an explicit image-grounding step, successfully retrieving correct images.
  \textcolor{okgreen}{Green borders} indicate images containing only the affirmed concept~\eqref{eq:correct}, while \textcolor{errred}{red borders} indicate confusing images containing the additional negated concept~\eqref{eq:confusing}. 
    }
  \label{fig:teaser}
\end{figure}

\vspace{-0.2cm}
\section{Preliminaries}
\vspace{-0.2cm}
\paragraph{Vision-language retrieval.} We consider a dual-encoder vision-language model~\citep{zhai2023sigmoid, radford2021learning, tschannen2025siglip}, consisting of a text encoder $f_{\mathcal{T}}\colon\mathcal{T} \to \mathbb{R}^{d}$ and an image encoder $f_\mathcal{I}\colon \mathcal{I} \to \mathbb{R}^d$, both producing $\ell_2$-normalized embeddings in a shared $d$-dimensional space.
Given a text query $q \in \mathcal{T}$ and a gallery $\{i_j\}_{j=1}^{M} \subset \mathcal{I}$, image retrieval ranks $M$ candidate images by their inner product to the query
\vspace{-0.1cm}
\begin{equation}
s(q, i) = f_{\mathcal{T}}(q)^\top f_{\mathcal{I}}(i).
\label{eq:score}
\end{equation}
%
\paragraph{Negation queries.}
We define a negation query as a compositional semantic constraint $q = (c^+, c^-)$, typically formulated as $ \text{``}c^+$~\textit{without}~$c^-\text{''}$ or equivalent exclusionary construction.
Here, $c^+$ denotes the \emph{affirmed} concept and $c^-$ denotes the \emph{negated} concept.
To evaluate the faithfulness of the retrieval system, we define a binary indicator function $\phi(i, c) \in \{0, 1 \}$ that denotes the presence of concept $c$ in image $i$. 
For a given $q = (c^+, c^-)$, we partition the gallery $\mathcal{I}$ into three mutually exclusive sets:

{\belowdisplayskip=-3pt
$\bullet$ \emph{Correct set} $~\mathcal{I}^+$: Images satisfying the affirmed concept while strictly excluding the negated concept
\begin{equation}
\label{eq:correct}
\mathcal{I}^{+} = \left\{i \in \mathcal{I} \;\middle|\; \phi(i,c^{+}) = 1,\; \phi(i,c^{-}) = 0 \right\}.
\end{equation}
}
\newline
{\belowdisplayskip=-3pt
$\bullet$ \emph{Confusing set} $~\mathcal{I}^{\pm}$: Images containing the affirmed concept but failing the negation constraint
\begin{equation}
\label{eq:confusing}
\mathcal{I}^{\pm} = \left\{i \in \mathcal{I} \;\middle|\; \phi(i,c^{+}) = 1,\; \phi(i,c^{-}) = 1 \right\}.   
\end{equation}
}
\newline
$\bullet$ \emph{Irrelevant set} $~\mathcal{I}^-$: Images that fail to satisfy the primary affirmed concept
\begin{equation}
\mathcal{I}^{-} = \left\{i \in \mathcal{I} \;\middle|\; \phi(i,c^{+}) = 0 \right\}.    
\end{equation}
In~\Cref{fig:teaser}, the affirmed concept $c^{+}$ is ``car'' and the negated concept $c^{-}$ is ``traffic light''. 
Faithful negation retrieval ranks $\mathcal{I}^+$ above $\mathcal{I}^{\pm}$, where $\mathcal{I}^{\pm}$ is the most challenging distractor.


\vspace{-0.2cm}
\section{Why single text embeddings cannot reject negation}
\label{sec:why-fail}
\vspace{-0.2cm}
Existing approaches only dilute the negated concept rather than reject it. We formalize this limitation by proving that contrastive training provides no gradient toward the negative coefficient that rejection requires. We confirm this limitation empirically on COCO-Confusion.

\vspace{-0.2cm}
\subsection{Existing approaches only dilute, not reject}
\vspace{-0.2cm}
Data-centric strategies typically augment training sets with negated captions and retrain the model to enhance its grasp of negation.
However, these approaches often yield marginal improvements. 
Despite the substantial computational overhead of retraining, the model's ability to distinguish between a caption and its negation remains far from satisfactory
~\citep{alhamoud2025vision, park2025know}.

To address this bottleneck, \citet{aggarwalseeing} introduced a \emph{training-free} embedding correction method that isolates the negated concept $c^-$ from the query $q$ and {project}s out its direction from the original query embedding
\begin{equation}
\tilde{\boldsymbol{t}} \;\propto\; f_{\mathcal{T}}(q) - \lambda \cdot\textsc{project}_{c^-} f_{\mathcal{T}}(q),
\label{eq:projection}
\end{equation}
where $\lambda$ is a scalar hyperparameter. 

\begin{figure}[t!]
 \centering
  \begin{tikzpicture}

   \node[font=\large\bfseries] at (6.5, 0.5) {Query: {``a photo of a {\textcolor{errred}{cat}} \emph{without} a \textcolor{ideblue}{dog}''}};

   \node[anchor=north] at (1.8, +0.2) {\includegraphics[width=4.2cm, height=3.2cm]{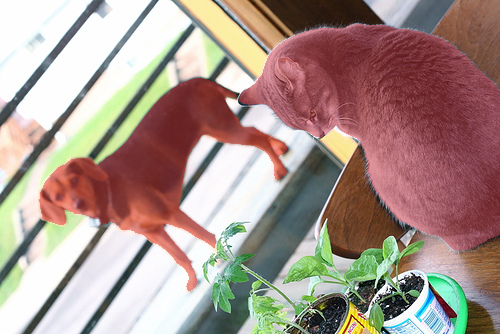}};
   \node[anchor=north] at (6.5, +0.2) {\includegraphics[width=4.2cm, height=3.2cm]{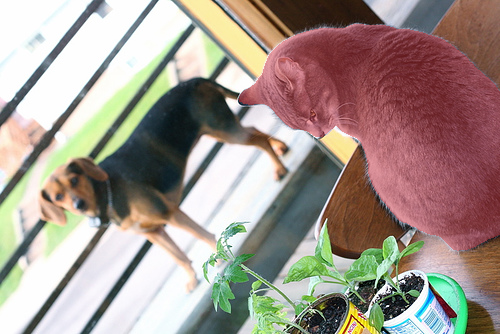}};
   \node[anchor=north] at (11.2, +0.2) {\includegraphics[width=4.2cm, height=3.2cm]{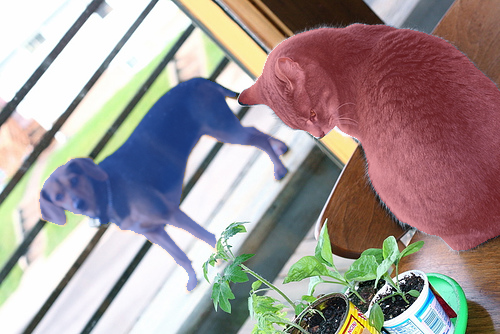}};

   \node[font=\bfseries, fill=errred!15, rounded corners=2pt, inner sep=4pt] at (1.8, -3.5) {\textcolor{errred}{Vanilla}};
   \node[font=\bfseries, fill=orange!15, rounded corners=2pt, inner sep=4pt] at (6.5, -3.5) {\textcolor{orange!80!black}{Embedding correction~\citep{aggarwalseeing}}};
   \node[font=\bfseries, fill=okgreen!15, rounded corners=2pt, inner sep=4pt] at (11.2, -3.5) {\textcolor{okgreen}{Proposed method}};

   \node[font=\small, text width=4.2cm, align=center] at (1.8, -4.3) {\textbf{Ignore}: \textcolor{errred}{Activates on both}\\ {Dog} negation ignored};
   \node[font=\small, text width=4.2cm, align=center] at (6.5, -4.3) {\textbf{Dilute}: Activates on \textcolor{errred}{cat} \\{Dog} feature suppressed};
   \node[font=\small, text width=4.2cm, align=center] at (11.2, -4.3) {\textbf{Reject}: Activates on \textcolor{errred}{cat}\\\textcolor{ideblue}{Dog explicitly penalized}};


  \end{tikzpicture}
  \vspace{-0.1cm}
  \caption{
  Given the query ``A photo of a {\textcolor{errred}{cat}} \emph{without} a \textcolor{ideblue}{dog}'', a {vanilla} VLM ignores the negation and activates on both concepts (left). Embedding correction approach~\citep{aggarwalseeing} removes the dog direction from the text embedding but never verifies the visual features, which amounts to passively diluting the concept (center). Faithful negation requires explicitly locating the dog in the image and penalizing its presence (right), which forms the crux of our method {\ourmethodlong}~(\ourmethodshort).}
  \label{fig:ignore-reject}
\end{figure}

While this approach yields notable empirical gains, 
we observe that the resulting embedding $\tilde{\boldsymbol{t}} \in \mathbb{R}^{d}$ remains \emph{agnostic} to the visual grounding of $c^{-}$.
The subtraction operation modifies the text embedding in isolation, failing to verify the absence of $c^-$ against its actual features grounded in the visual manifold.
Consequently, such embedding correction merely \emph{dilutes} the semantic influence of $c^{-}$ without penalizing its material presence in the image (see~\Cref{fig:ignore-reject}). 

We posit that faithful negation handling requires an \emph{explicit mechanism of rejection}---a capability that is structurally precluded by the conventional single embedding format.
This observation forms the crux of our argument: the conventional single-embedding VLMs lack the expressive bandwidth to encode exclusionary logic {within the same representation used for semantic matching}.

\vspace{-0.2cm}
\subsection{A structural barrier in single-embedding negation handling}
\label{sec:why-fail-theory}
\vspace{-0.2cm}

Under standard $\ell_2$-normalized contrastive training, the contrastive objective provides no learning signal toward the negative coefficient that rejection requires. Moreover, query-side post-hoc correction is also bounded by dilution. We formalize both limits below.
\vspace{-0.3cm}
\paragraph{Setup.}
For each concept $c$, let ${\boldsymbol{e}}_c^{*} \in \mathbb{R}^d$ denote a unit-norm proxy vector, ideally representing the concept $c$ in the space shared by both encoders. 
We define
\begin{equation}
\alpha_c(i) := f_{\mathcal{I}}(i)^\top {\boldsymbol{e}}_c^{*} 
\quad (\text{image activation}), \qquad
\beta_c(q) := f_{\mathcal{T}}(q)^\top {\boldsymbol{e}}_c^{*} 
\quad (\text{text coefficient}).
\label{eq:alpha_beta}
\end{equation}
For analytical tractability we adopt a linear concept model: $f_{\mathcal{I}}(i) = \sum_c \alpha_c(i) {\boldsymbol{e}}_c^{*}$ on an orthonormal concept basis. We focus on negation queries $q = $ ``$c^+$ \emph{without} $c^-$''.
\begin{assumption}[Concept Cone, idealized]
\label{asm:cone}
In the idealized model, concept activations are non-negative: $\alpha_c(i) \geq 0$, with $\alpha_c(i) > 0$ iff $c \in i$. This captures the intuition that semantic similarity measures presence, not anti-presence.
\end{assumption}
\begin{lemma}[Rejection criterion]
\label{lem:rejection-criterion}
For $q = $ ``$c^+$ without $c^-$'', consider images 
$i^+ \in \mathcal{I}^+$ and $i^\pm \in \mathcal{I}^\pm$ that 
differ only along the $c^-$ axis. Under the linear concept model,
\begin{equation}
s(i^+, q) - s(i^\pm, q) = -\alpha_{c^-}(i^\pm)\,\beta_{c^-}(q).
\end{equation}
Faithful retrieval demands $s(i^+, q) > s(i^\pm, q)$, and since 
$\alpha_{c^-}(i^\pm) > 0$, this is equivalent to
\begin{equation}
\boxed{\beta_{c^-}(q) < 0.}
\label{eq:rejection-condition}
\end{equation}
\end{lemma}

\begin{theorem}[In-kind alignment carries no gradient toward $\beta_{c^-}(q) < 0$]
\label{thm:beta-zero}
Consider the positive-alignment objective
\begin{equation}
\mathcal{A}(q) = f_{\mathcal{T}}(q)^\top \boldsymbol{\mu}(q), 
\qquad \boldsymbol{\mu}(q) := \mathbb{E}_{i \in \mathcal{I}_q}
[f_{\mathcal{I}}(i)],
\end{equation}
the positive-pair component of standard contrastive losses. For 
$q = $ ``$c^+$ without $c^-$'', every $i \in \mathcal{I}_q$ satisfies 
$c^- \notin i$, so $\mu_{c^-}(q) = 0$, and decomposing 
$\mathcal{A}(q) = \sum_k \beta_k(q)\,\mu_k(q)$ gives
\begin{equation}
\boxed{\;\frac{\partial \mathcal{A}(q)}{\partial \beta_{c^-}(q)} 
\;\equiv\; 0.\;}
\label{eq:gradient-zero}
\end{equation}
The objective is invariant in $\beta_{c^-}(q)$: positive-alignment training provides no gradient that could drive $\beta_{c^-}(q)$ below zero.
\end{theorem}
\vspace{-0.1cm}
\begin{corollary}[Optimal text coefficient under unit-norm constraint]
\label{cor:beta-star-zero}
Under the unit-norm constraint $\|f_{\mathcal{T}}(q)\| = 1$ standard for 
$\ell_2$-normalized encoders, the maximizer of $\mathcal{A}(q) = 
\beta(q)^\top \boldsymbol{\mu}(q)$ is, by Cauchy--Schwarz, 
$\beta^*(q) = \boldsymbol{\mu}(q)/\|\boldsymbol{\mu}(q)\|$. Since 
$\mu_{c^-}(q) = 0$ for negation queries, $\beta^*_{c^-}(q) = 0$. 
Combined with Lemma~\ref{lem:rejection-criterion}, this implies 
$s(i^+, q) = s(i^\pm, q)$ at the optimum: even an optimally-trained 
encoder under in-kind alignment cannot rank correct above confusing.
(Details in Appendix~\ref{app:theory}.)
\end{corollary}
\vspace{-0.3cm}
\paragraph{Beyond in-kind alignment.}
Hard-negative supervision lies outside the positive-alignment regime but, under token frequency asymmetry, collapses to a generic negation offset rather than concept-specific rejection (Appendix~\ref{app:theory}). Post-hoc embedding correction still only reaches $\tilde{\beta}_{c^-}(q) = 0$ and fails to overcome the Lemma~\ref{lem:rejection-criterion} barrier. $\tilde{\beta}_{c^-}(q) = 0$ is the ceiling of all existing approaches, training-based or post-hoc, yet remains insufficient for rejection.
\vspace{-0.2cm}
\subsection{The limit of dilution-based approaches}
\begin{figure}[t]
\centering
\small
\begin{tabular}{@{}cc@{}}
\begin{minipage}[c]{0.45\linewidth}
\centering
\includegraphics[width=0.8\linewidth]{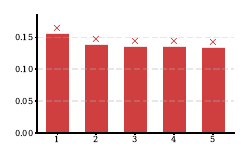}\\[-10pt]
\hspace{1.5em}\textbf{Raw SigLIP2}
\end{minipage}
&
\begin{minipage}[c]{0.45\linewidth}
\centering
\includegraphics[width=0.8\linewidth]{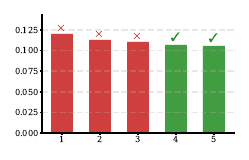}\\[-10pt]
\hspace{1.5em}\textbf{Embedding Correction}
\end{minipage}
\end{tabular}
\caption{\textbf{Dilution cannot separate correct from confusing images.} Top-5 similarity scores for \textit{``a car is present but no traffic light''} on COCO-Confusion, SigLIP2 SO400M. \textbf{Raw SigLIP2} (left): all top-5 are confusing (red) with uniform scores ($\approx 0.15$).
\textbf{Embedding Correction} (right): scores drop overall ($\approx 0.12$), but red (confusing) and green (correct) bars remain interleaved, failing to separate correct from confusing.}
\label{fig:dilution-failure}
\end{figure}

\vspace{-0.2cm}
We demonstrate these limitations with an example query on COCO-Confusion (Figure~\ref{fig:dilution-failure}).
%
%
%
\definecolor{okgreen}{RGB}{34,139,34}
\definecolor{errred}{RGB}{200,30,30}
\vspace{-0.2cm}
\paragraph{Vanilla VLM.}
Vanilla VLM does not process the negation word ``no,'' so the query is effectively treated as \emph{``a car is present and traffic light.''}
As a result, images containing both car and traffic light receive the highest similarity score---all top-5 results are confusing images (red bar) with nearly uniform scores (${\approx}\, 0.15$).
\vspace{-0.2cm}
\paragraph{Embedding correction.}
Embedding Correction dilutes the traffic light concept by removing its direction from the text embedding.
As a result, the overall scores of confusing images decrease (${\approx}\, 0.12$).
However, since Embedding Correction \emph{ignores} traffic light rather than \emph{rejecting} it, images containing both car and traffic light still receive high similarity scores. The score bars remain nearly uniform across confusing (red) and correct (green) images, failing to separate the two groups.

This confirms the dilution ceiling: single-embedding scoring cannot reject $c^-$. \textbf{Negation needs multi-dimensional scoring}. One dimension to affirm $c^+$, another to penalize the image-side activation of $c^-$. 
\ourmethodshort\ realizes this decomposition.

\vspace{-0.2cm}
\section{Proposed method: \ourmethodlong}
\label{sec:proposed-method}
\vspace{-0.2cm}
Section~\ref{sec:why-fail} established that single-embedding scoring cannot achieve rejection: contrastive training provides no gradient toward $\beta_{c^-}(q) < 0$ (Theorem~\ref{thm:beta-zero}), and post-hoc correction only reaches $\tilde{\beta}_{c^-}(q) = 0$, still short of the rejection condition required by Lemma~\ref{lem:rejection-criterion}. All single-embedding approaches share this limitation.

We propose \textbf{\ourmethodlong~(\ourmethodshort)}, a training-free method that extends scoring from one dimension to two: \emph{(i)}~a \emph{text-side affirmation dimension} that handles $c^+$ via a corrected text embedding $\tilde{\boldsymbol{t}}$, and \emph{(ii)}~an \emph{image-side rejection dimension} that handles $c^-$ by detecting and penalizing its visual presence in $i$.
The full scoring is
\begin{equation}
s(q, i) \;=\; \underbrace{\tilde{\boldsymbol{t}}^\top f_{\mathcal{I}}(i)}_{\text{Dim 1: Affirm } c^+} \;-\;\; w_n\cdot\underbrace{p_{c^-}(i)}_{\text{Dim 2: Reject } c^-},
\label{eq:hard no}
\end{equation}
$p_{c^-}(i)$ is the relative grounding score defined in Eq.~\eqref{eq:detection}, and $w_n$ is a scaling parameter.
Dimension~1 operates on the text side and is constrained by Theorem~\ref{thm:beta-zero}---by itself, it can only achieve dilution.
Dimension~2 operates on the image side using a separate computation that does not pass through $f_{\mathcal{T}}(\cdot)$, hence is not bound by Theorem~\ref{thm:beta-zero}'s unit-norm budget.
The two dimensions share no parameter budget, allowing $p_{c^-}(i)$ to provide the rejection signal that Lemma~\ref{lem:rejection-criterion} requires and Corollary~\ref{cor:beta-star-zero} forbids in any single-dimension formulation.
\vspace{-0.2cm}
\subsection{Text-side affirmation}
\label{sec:text-affirmation}
\vspace{-0.2cm}
Given a natural-language query $q$ such as ``a cat without a dog'', we first extract the negated concept $c^-$ and the affirmed concepts $\mathcal{C}^+$ via spaCy's dependency parser~\citep{honnibal2020spacy}.
Negation cues---\emph{no, not, without, absent, missing, free of}---are identified by dependency relation (\texttt{neg}) or lemma matching, and the head noun of the negation span yields $c^-$; remaining noun phrases form $\mathcal{C}^+$.

We then compute the corrected text embedding by projecting out the $c^-$ direction:
\begin{equation}
\tilde{\boldsymbol{t}} \;=\; \textsc{normalize}\!\left(f_{\mathcal{T}}(q) - \textsc{project}_{c^-} f_{\mathcal{T}}(q)\right)
\label{eq:text-correction}
\end{equation}
where $\mathrm{proj}_{c^-}(\cdot)$ denotes orthogonal projection onto the $c^-$ direction, making $\tilde{\boldsymbol{t}}$ orthogonal to $\boldsymbol{e}_{c^-}$. This achieves $\beta_{c^-}(\tilde{\boldsymbol{t}}) = 0$, matching the constrained optimum of Corollary~\ref{cor:beta-star-zero} --- yet by Lemma~\ref{lem:rejection-criterion} still insufficient for rejection.

\vspace{-0.2cm}
\subsection{Image-side rejection}
\vspace{-0.2cm}

The second scoring dimension produces an image-conditioned penalty that grows when $c^-$ is grounded in image $i$ and vanishes when it is not.
A naive thresholding on $f_{\mathcal{I}}(i)^\top \mathbf{e}_{c^-}$ fails because raw similarity scales vary substantially across images: some images yield globally high similarity to all concepts, producing false positives.

We then compute a relative detection score $p_{c^-}(i)$ using a per-image reference $\tau(i)$
\begin{equation}
p_{c^-}(i) \;=\; \sigma\!\left( \kappa\,\big(f_{\mathcal{I}}(i)^\top \boldsymbol{e}_{c^-} - \tau(i)\big)\right) - \frac{1}{2}\sigma\!\left( \kappa\,\big(f_{\mathcal{I}}(i)^\top \boldsymbol{e}_{c^+} - \tau(i)\big)\right),
\label{eq:detection}
\end{equation}
\begin{equation}
\tau(i) \;=\; \frac{1}{|\mathcal{C}^+|}\sum_{c \in \mathcal{C}^+} f_{\mathcal{I}}(i)^\top \boldsymbol{e}_c.
\label{eq:tau}
\end{equation}
where $\sigma(\cdot)$ is the sigmoid acting as a soft thresholding operator --- penalizing $c^-$ only when $\alpha_{c^-}(i)$ exceeds $\tau(i)$ --- and $\kappa$ controls the sharpness of the transition.
Real VLMs deviate from the ideal encoder of Assumption~\ref{asm:cone}: 
$\alpha_{c^-}(i)$ remains non-zero even when $c^-$ is absent. We 
therefore anchor the threshold to the image's own affirmed-concept 
similarity, $\tau(i) = \frac{1}{|\mathcal{C}^+|}\sum_{c \in \mathcal{C}^+} 
\alpha_c(i)$, providing a query-relevant, image-specific baseline. 
When $c^-$'s similarity exceeds this baseline, the first term drives $p_{c^-}(i) \to 1$, and 
otherwise it falls to $0$.
The second term further suppresses false positives that arise when $c^+$ and $c^-$ are semantically correlated (e.g., \emph{cat} and \emph{dog}). In such cases $\alpha_{c^-}(i)$ may exceed $\tau(i)$ even when $c^-$ is absent, and discounting by the $c^+$-side activation prevents correct images from spurious penalty. 
Crucially, $p_{c^-}(i)$ is computed entirely from $f_{\mathcal{I}}(i)$ and concept directions---it does \textbf{not} pass through the text-query embedding $f_{\mathcal{T}}(q)$, which is what allows the rejection dimension to penalize $c^-$ regardless of the alignment-induced ceiling on $\beta_{c^-}^*$.

\vspace{-0.2cm}
\subsubsection{Why \ourmethodshort~avoids the constraint}
\label{sec:escape}
\vspace{-0.2cm}

Corollary~\ref{cor:beta-star-zero} bounds the contribution of any single text-side coordinate: under the unit-norm constraint, $\beta_{c^-}^* = 0$ exactly.
Dimension~1 respects this bound---$\tilde{\boldsymbol{t}}^\top f_{\mathcal{I}}(i)$ is a single-vector dot product and inherits the $\beta_{c^-}^* = 0$ ceiling.

Dimension~2 sidesteps the bound by introducing a separate scoring axis computed outside the text-side unit-sphere budget.
The total score is no longer a single dot product; it is a two-dimensional composition where each dimension operates in its own budget.
The matched-pair test of Lemma~\ref{lem:rejection-criterion}---which requires a negative ranking signal---is satisfied by Dimension~2.

The asymmetry of natural visual data---\emph{presence learns; absence does not}---is resolved by decomposing the problem into two dimensions: Dimension~1 handles affirmation, where the data measure naturally provides positive signal ($\mu_{c^+} > 0$); Dimension~2 handles rejection, where the data measure provides no signal ($\mu_{c^-} = 0$) but direct image-side detection circumvents this gap entirely.
\vspace{-0.2cm}
\section{Experiments}
\label{sec:experiments}
\vspace{-0.2cm}
\subsection{Introducing COCO-Confusion: A benchmark for negation rejection}
Existing negation retrieval benchmarks such as NegBench evaluate retrieval over a large, unfiltered gallery (e.g., 5{,}000 COCO~\citep{lin2014microsoft} images), where the negated concept rarely co-occurs with the affirmed concept in the same image. In such settings, even simple embedding correction methods perform well because there are few confusing images to distinguish from correct ones. This means existing benchmarks cannot distinguish between \emph{ignoring} and \emph{rejecting} the negated concept.
To expose this critical gap, we construct COCO-Confusion, a retrieval benchmark specifically designed to test whether a method can reject confusing images that contain the negated concept.
\vspace{-0.3cm}
\paragraph{Construction.}
We build COCO-Confusion from COCO val2017 using bounding box annotations.
For each concept pair $(c^+, c^-)$, we partition the annotated images into two groups: \emph{correct} images containing $c^+$ but not $c^-$, and \emph{confusing} images containing both $c^+$ and $c^-$.
We consider two settings based on query complexity: COCO-Confusion-2C uses one affirmed and one negated concept (e.g., ``A car is present but no traffic light''; 950 pairs), and COCO-Confusion-3C uses two affirmed concepts and one negated concept (e.g., ``An apple and banana are present but no bowl''; 1{,}740 pairs).
Each group is capped at 50 images per pair to ensure balanced galleries.
We verify all images using OWLv2~\citep{minderer2023scaling}, an open-vocabulary object detector, to filter annotation noise where the negated concept is heavily occluded, truncated, or visually ambiguous in the bounding-box annotations, removing pairs where the detector cannot confirm concept presence. (details in Appendix~\ref{app:coco_conf})
The final benchmark contains 2{,}690 verified pairs across 4{,}708 unique images.
\vspace{-0.3cm}
\paragraph{Evaluation metric.}
For each pair, the gallery consists of its correct and confusing images. We report mean Average Precision (mAP), which measures how well a method ranks correct images above confusing ones. High mAP indicates successful rejection of confusing images beyond dilution.
\vspace{-0.3cm}
\paragraph{Why COCO-Confusion matters.}
On existing retrieval benchmarks, Embedding Correction method achieves near-optimal performance because confusing images are rare in the gallery.
On COCO-Confusion, every gallery is designed to contain confusing images, making it the first benchmark that explicitly measures negation \emph{rejection} rather than \emph{dilution}.
\vspace{-0.2cm}
\subsection{Experimental setup}
\vspace{-0.2cm}
\paragraph{Backbones.}
We evaluate on 9 pretrained VLMs spanning three model families and diverse scales:
CLIP (ViT-B/32, L/14, H/14~\citep{dosovitskiy2020image}; LAION-2B~\citep{schuhmann2022laion}), EVA-02~\citep{fang2024eva} (ViT-B/16, L/14; Merged-2B), and SigLIP/SigLIP2 (ViT-B/16, B/32, L/16, SO400M; WebLI with sigmoid loss). All weights are public and frozen.
\vspace{-0.2cm}
\paragraph{Evaluation tasks.}
We evaluate on three complementary tasks.
(1)~\textbf{MCQ:} Given an image and four caption options, select the correct one.
We report accuracy on COCO MCQ and VOC2007 MCQ from NegBench. This measures negation \emph{comprehension}. (2)~\textbf{Standard negation retrieval:} Given a negated query, retrieve matching images from a large gallery. 
We report Recall@5 on COCO and MSRvtt~\citep{xu2016msr} negated retrieval from NegBench. This measures negation handling on natural galleries where confusing images are rare. (3)~\textbf{COCO-Confusion retrieval:} Given a negated query, retrieve correct images from a gallery containing both correct and confusing images. We report mAP on COCO-Confusion-2C and COCO-Confusion-3C setting. This measures negation \emph{rejection}.
Qualitative retrieval examples are provided in Appendix.
\vspace{-0.3cm}
\paragraph{Baselines.}
We compare against Raw VLM, which uses standard cosine similarity with no negation processing, and Embedding Correction, a training-free method that removes the negated concept direction from the text embedding. Since Embedding Correction reports results only for CLIP ViT-B/32 and does not release code, we use their reported numbers for this backbone. For CLIP ViT-B/32, we additionally compare against NegCLIP and NegationCLIP.

\begin{table*}[t]
\centering
\footnotesize
\caption{Negation performance on CLIP ViT-B/32~\citep{radford2021learning}. We compare against training-based negation methods~\citep{yuksekgonul2022and, park2025know} and training-free embedding correction method (EC)~\cite{aggarwalseeing} across MCQ, Retrieval, and our COCO-Confusion benchmark. (*) indicates models fine-tuned on CC12M-NegFull~\citep{alhamoud2025vision}.
Best in \textbf{bold}; second-best \underline{underlined}.}
\setlength{\tabcolsep}{3pt}
\begin{tabular}{l c c c c c c c c c}
\toprule
 & \multicolumn{3}{c}{\textbf{MCQ}} & \multicolumn{6}{c}{\textbf{Retrieval}} \\
\cmidrule(lr){2-4} \cmidrule(lr){5-10}
 & \textbf{COCO} & \textbf{VOC2007} & \textbf{MSRvtt} & \multicolumn{2}{c}{\textbf{COCO}} & \multicolumn{2}{c}{\textbf{MSRvtt}} & \textbf{Conf-2C} & \textbf{Conf-3C} \\
\midrule
\textbf{Model} & & & & R@5 & R-Neg@5 & R@5 & R-Neg@5 & mAP & mAP \\
\midrule
CLIP~\citep{radford2021learning} & 24.4 & 27.2 & 30.1 & \underline{65.3} & 57.9 & \underline{55.7} & 48.8 & 63.2 & 60.5 \\
\midrule
NegCLIP~\citep{yuksekgonul2022and} & 28.5 & 30.4 & 28.4 & \textbf{69.5} & \textbf{65.2} & \textbf{57.0} & \underline{51.6} & 64.4 & 61.4 \\
CLIP* (NegBench)~\citep{alhamoud2025vision} & 54.4 & 54.9 & 45.3 & 54.9 & 52.3 & 48.8 & 43.1 & 68.8 & 65.0 \\
NegationCLIP~\citep{park2025know} & 35.9 & 44.2 & 37.4 & 62.8 & \underline{62.0} & 52.5 & 49.7 & 72.9 & 68.4 \\
\midrule
EC~\citep{aggarwalseeing} & 70.0 & 75.5 & 51.6 & {65.2} & 58.3 & \underline{55.7} & 51.0 & \underline{73.2} & \underline{70.8} \\
\textbf{KTN (Ours)} & \textbf{79.5} & \textbf{92.1} & \textbf{66.9} & {65.2} & \underline{62.0} & \underline{55.7} & \textbf{51.9} & \textbf{82.8} & \textbf{75.2} \\
\bottomrule
\end{tabular}
\label{tab:main}
\end{table*}
\vspace{-0.1cm}
\begin{table*}[h]
\centering
\footnotesize
\caption{Negation performance across multiple VLM backbones~\citep{zhai2023sigmoid,radford2021learning,tschannen2025siglip,fang2024eva}. Each backbone shows three rows: Vanilla VLM, embedding correction (EC)~\cite{aggarwalseeing}, and our \ourmethodshort. \textbf{Bold} indicates best-per-metric across the three methods. \ourmethodshort\ consistently outperforms both vanilla and embedding correction on MCQ comprehension and Confusion retrieval, while preserving standard retrieval R@5.}
\setlength{\tabcolsep}{3pt}
\begin{tabular}{l c c c c c c c c c c}
\toprule
 & & \multicolumn{3}{c}{\textbf{MCQ}} & \multicolumn{6}{c}{\textbf{Retrieval}} \\
\cmidrule(lr){3-5} \cmidrule(lr){6-11}
 & & \textbf{COCO} & \textbf{VOC2007} & \textbf{MSRvtt} & \multicolumn{2}{c}{\textbf{COCO}} & \multicolumn{2}{c}{\textbf{MSRvtt}} & \textbf{Conf-2C} & \textbf{Conf-3C}\\
\midrule
\textbf{Model} & \textbf{Backbone} & & & & R@5 & R-Neg@5 & R@5 & R-Neg@5 & mAP & mAP \\
\midrule
Vanilla & \multirow{3}{*}{\shortstack{CLIP~\citep{radford2021learning}\\ViT-L/14}}  & 25.4 & 25.3 & 30.9 & 70.8 & 64.1 & 60.4 & 55.0 & 62.1 & 59.4 \\
EC~\citep{aggarwalseeing} & & 48.4 & 45.3 & 41.9 & 70.8 & 66.6 & 60.4 & 55.9 & 73.8 & 67.4 \\
\textbf{\ourmethodshort\ (Ours)} & & \textbf{79.6} & \textbf{92.4} & \textbf{69.1} & 70.8 & \textbf{67.8} & \textbf{60.5} & \textbf{56.5} & \textbf{82.4} & \textbf{74.8} \\
\midrule
Vanilla & \multirow{3}{*}{\shortstack{CLIP~\citep{radford2021learning}\\ViT-H/14}}  & 30.2 & 31.9 & 29.9 & 73.2 & 66.8 & 62.6 & 55.0 & 62.0 & 59.3 \\
EC~\citep{aggarwalseeing} & & 53.9 & 49.9 & 43.3 & 73.2 & 68.5 & 62.6 & 57.8 & 72.2 & 66.1 \\
\textbf{\textbf{\ourmethodshort\ (Ours)}} & & \textbf{79.4} & \textbf{92.2} & \textbf{68.1} & 73.2 & \textbf{69.9} & 62.6 & \textbf{59.2} & \textbf{81.6} & \textbf{74.0} \\
\midrule
Vanilla & \multirow{3}{*}{\shortstack{EVA-02~\citep{fang2024eva}\\ViT-B/16}}  & 30.2 & 27.8 & 29.2 & 66.6 & 61.2 & 58.7 & 51.1 & 62.3 & 58.7 \\
EC~\citep{aggarwalseeing} & & 44.3 & 40.0 & 43.5 & 66.6 & 63.5 & 58.7 & \textbf{53.8} & 76.1 & 68.5 \\
\textbf{\textbf{\ourmethodshort\ (Ours)}} & & \textbf{77.1} & \textbf{91.2} & \textbf{63.8} & 66.6 & \textbf{64.4} & 58.7 & 53.5 & \textbf{84.4} & \textbf{76.2} \\
\midrule
Vanilla & \multirow{3}{*}{\shortstack{EVA-02~\citep{fang2024eva}\\ViT-L/14}}  & 31.5 & 31.9 & 30.4 & 70.9 & 65.5 & 62.8 & 56.9 & 61.1 & 57.9 \\
EC~\citep{aggarwalseeing} & & 38.3 & 25.4 & 39.7 & 70.9 & 67.7 & 62.8 & 58.2 & 77.4 & 68.1 \\
\textbf{\textbf{\ourmethodshort\ (Ours)}} & & \textbf{79.7} & \textbf{92.4} & \textbf{65.9} & 70.9 & \textbf{68.5} & 62.8 & \textbf{58.4} & \textbf{84.6} & \textbf{76.2} \\
\midrule
Vanilla & \multirow{3}{*}{\shortstack{SigLIP~\citep{zhai2023sigmoid}\\ViT-B/16}}  & 28.8 & 31.1 & 32.2 & 72.3 & 65.0 & 56.1 & 49.7 & 62.3 & 60.2 \\
EC~\citep{aggarwalseeing} & & 56.3 & 77.1 & 51.6 & 72.3 & 68.6 & 56.1 & 50.5 & 79.2 & 72.2 \\
\textbf{\textbf{\ourmethodshort\ (Ours)}} & & \textbf{71.1} & \textbf{82.9} & \textbf{59.2} & 72.3 & \textbf{69.6} & 56.1 & \textbf{51.5} & \textbf{85.5} & \textbf{78.1} \\
\midrule
Vanilla & \multirow{3}{*}{\shortstack{SigLIP2~\citep{tschannen2025siglip}\\ViT-B/32}}  & 27.1 & 27.3 & 28.2 & 73.2 & 64.8 & 56.9 & \textbf{51.3} & 61.8 & 60.2 \\
EC~\citep{aggarwalseeing} & & 63.3 & 78.2 & 51.2 & 73.2 & 68.4 & 56.9 & 49.2 & 81.4 & 74.4 \\
\textbf{\textbf{\ourmethodshort\ (Ours)}} & & \textbf{78.5} & \textbf{93.4} & \textbf{61.4} & 73.2 & \textbf{69.5} & 56.9 & 50.5 & \textbf{86.7} & \textbf{79.2} \\
\midrule
Vanilla & \multirow{3}{*}{\shortstack{SigLIP2~\citep{tschannen2025siglip}\\ViT-L/16}}  & 25.2 & 24.1 & 28.8 & 77.1 & 71.0 & 64.1 & 56.8 & 61.1 & 59.1 \\
EC~\citep{aggarwalseeing} & & 57.7 & 71.8 & 53.9 & 77.1 & 74.0 & 64.1 & 58.0 & 78.6 & 71.4 \\
\textbf{\textbf{\ourmethodshort\ (Ours)}} & & \textbf{80.5} & \textbf{94.6} & \textbf{64.6} & 77.1 & \textbf{75.3} & 64.1 & \textbf{58.4} & \textbf{87.0} & \textbf{79.6} \\
\midrule
Vanilla & \multirow{3}{*}{\shortstack{SigLIP2~\citep{tschannen2025siglip}\\So400m}}    & 23.9 & 23.8 & 28.9 & \textbf{78.0} & 71.5 & 64.9 & 58.0 & 61.5 & 59.8 \\
EC~\citep{aggarwalseeing} & & 55.1 & 71.1 & 54.0 & \textbf{78.0} & 74.6 & 64.9 & 59.5 & 78.6 & 69.8 \\
\textbf{\textbf{\ourmethodshort\ (Ours)}} & & \textbf{81.0} & \textbf{94.6} & \textbf{65.7} & 77.9 & \textbf{75.9} & 64.9 & \textbf{59.6} & \textbf{87.1} & \textbf{79.3} \\
\bottomrule
\end{tabular}
\label{tab:multi-backbone-app}
\end{table*}
\vspace{-0.2cm}
\subsection{Main results}
\vspace{-0.1cm}
\paragraph{MCQ.}
Our method substantially improves negation comprehension 
across all MCQ benchmarks. On CLIP ViT-B/32 
(Table~\ref{tab:main}), we achieve 79.5\% on COCO, 
92.1\% on VOC2007, and 66.9\% on MSRvtt, improving over 
Embedding Correction by +9.5, +16.6, and +15.3 points 
respectively. Both training-based baselines fall far 
behind: NegCLIP and NegationCLIP 
show only marginal gains over the raw 
baseline despite requiring retraining. These improvements 
generalize consistently across all backbones 
(Table~\ref{tab:multi-backbone-app}): MCQ accuracy improves by 
+42 to +57 points on COCO, +51 to +71 points on VOC2007, 
and +31 to +37 points on MSRvtt relative to the raw 
baseline. The gains are consistent regardless of model 
family (CLIP, EVA-02, SigLIP, SigLIP2) or scale 
(ViT-B to SO400M).
\vspace{-0.3cm}
\paragraph{Retrieval.}
On standard negation retrieval, where confusing images 
are rare in the gallery, our method preserves R@5 and 
consistently improves R-Neg@5. On CLIP ViT-B/32, R@5 
remains at 65.2 while R-Neg@5 improves from 58.3 to 
62.0 (+3.7). Across all backbones, R-Neg@5 gains range 
from +1.5 to +4.7 on COCO and +1.5 to +4.2 on MSRvtt, 
with no degradation in R@5. 
\vspace{-0.3cm}
\paragraph{COCO-Confusion.}
The largest gains appear on COCO-Confusion, the benchmark 
designed to test negation rejection rather than dilution. 
On CLIP ViT-B/32, our method achieves 82.8\% mAP on 
Conf-2C and 75.2\% on Conf-3C, improving over Embedding 
Correction by +9.6 and +4.4 points. Across all backbones, 
the pattern is striking: raw baselines cluster around 
61--65\% mAP regardless of model scale, confirming that 
larger models do not resolve negation failure. Embedding 
Correction provides only modest improvement. Our method 
consistently reaches 81--87\% on Conf-2C and 74--79\% on 
Conf-3C, with the best result of 87.1\% mAP on SigLIP2 
SO400M. This confirms that the image-side rejection 
dimension successfully separates correct from confusing 
images, precisely the gap that single-embedding scoring 
cannot close.

\vspace{-0.2cm}
\subsection{Ablation study}
\vspace{-0.1cm}
\begin{table}[t]
\centering
\small
\caption{\textbf{Ablation study} (SigLIP2 SO400M). Dimension 1 alone (Proj) achieves dilution. Dimension 2 provides rejection.}
\label{tab:ablation}
\begin{tabular}{lccc}
\toprule
Method & COCO MCQ & R-Neg@5 & Conf-2C \\
\midrule
SigLIP2 & 23.9 & 71.5 & 61.5 \\
+ Proj (Dim 1) & 55.1 & 74.6 & 78.6 \\
\textbf{+ Ours (Dim 1+2)} & \textbf{81.0} & \textbf{75.9} & \textbf{87.1} \\
\bottomrule
\end{tabular}
\end{table}
Table~\ref{tab:ablation} isolates the contribution of each scoring dimension on SigLIP2 SO400M.
Dimension~1 alone (Proj) already yields a large gain over Raw on MCQ (+31.2pp) and COCO-Confusion (+17.1pp), confirming that dilution is effective when the task requires distinguishing affirmative from negated captions.
However, adding Dimension~2 \ourmethodshort\ produces a further +25.9pp on MCQ and +8.5pp on COCO-Confusion, gains that are inaccessible to any single-embedding method by Theorem~\ref{thm:beta-zero}.
On standard negation retrieval (R-Neg@5), the gap is small (+1.3pp), consistent with the observation that confusing images are rare in unfiltered galleries and dilution alone suffices.
The ablation confirms that the dominant source of improvement is the image-side rejection dimension, precisely the signal that the dilution ceiling forbids.
\vspace{-0.2cm}
\section{Related work}
\vspace{-0.2cm}
%
%
\paragraph{Negation failure in VLMs.}
A growing body of work documents that VLMs systematically fail on negation. The ARO benchmark~\citep{yuksekgonul2022and} showed that CLIP behaves as a bag-of-words and largely ignores word order and negation. NegBench~\citep{alhamoud2025vision} evaluated multiple VLMs across 18 negation tasks and found near-chance performance, with CLIP scoring 82\% on affirmative queries but only 3\% on negated ones for VOC2007. NegationCLIP~\citep{park2025know} traced this failure to a training data shortfall and reported that only 0.7\% of LAION-400M captions contain any negation word. DCSM~\citep{kang2025clip} provides a theoretical account and argues that no CLIP-like joint embedding space can support both basic content representation and negation simultaneously.
\vspace{-0.2cm}
\paragraph{Approaches for improving negations awareness.}
Data-centric methods augment training data with negation captions and retrain the model. ConCLIP~\citep{singh2024learn} constructed 228K negation triples and fine-tuned with negated-caption distractors. NegationCLIP~\citep{park2025know} generated visually grounded negation captions through LLM and MLLM pipelines. However, these approaches require retraining and show limited generalization. On the training-free side, the embedding correction method~\citep{aggarwalseeing} showed that CLIP text embeddings follow compositional arithmetic and proposed a rule-based projection that extracts negated concepts and removes their semantic direction from the  embedding. This is effective at reducing negation bias but limited to dilution rather than rejection. We go beyond this limitation by adding a separate dimension that actively penalizes the negated concept's visual presence.

\vspace{-0.2cm}
\section{Conclusion}
\vspace{-0.2cm}

Representing negation within a single embedding vector 
is fundamentally limited. Even with perfect training 
data and infinite capacity, the non-negative nature of 
visual concept activations drives the text coefficient 
on negated concepts to zero, making dilution the 
theoretical ceiling of single-embedding scoring. We 
proposed \ourmethodlong~(\ourmethodshort), a training-free multi-dimensional scoring method that breaks this ceiling by separating affirmation and rejection into independent dimensions. 
Combined with COCO-Confusion, a benchmark designed to 
measure rejection beyond dilution, experiments 
across nine backbones confirm consistent improvements 
in MCQ, negation retrieval, and confusion rejection.
Beyond performance gains, COCO-Confusion makes the distinction between dilution and rejection measurable for the first time, revealing a gap that prior benchmarks could not expose.

\vspace{-0.2cm}
\section*{Limitations}
\ourmethodshort's image-side rejection relies on cosine similarity, which can yield high scores for $c^-$ even when it is absent from the image. We mitigate this with adaptive thresholding using a per-image reference $\tau(i)$ (Eq.~\ref{eq:tau}), but it does not fully eliminate false positives. 

Our method targets dual-encoder VLMs (CLIP, SigLIP) that score via a single dot product. Extending to generative VLMs like LLaVA~\citep{liu2023visual, liu2024improved} is a natural next step. Their scoring is conditioned on autoregressive decoding rather than embedding similarity, so the text coefficient $\beta_{c^-}$ analyzed in our impossibility result does not directly apply. Still, the broader principle that rejection requires a scoring signal beyond a single embedding axis likely transfers, and we expect LLaVA-style architectures to also benefit from a multi-dimensional formulation that rejects the negated concept rather than merely diluting it.

\newpage
\bibliographystyle{IEEEtranN}
\bibliography{abrv,ref}

\newpage
\newpage
\appendix

\section{Theoretical details}
\label{app:theory}

\subsection{Per-image budget compression}
\label{app:budget}

For each image $i$, let $\mathcal{P}(i) := \{c : \alpha_c(i) > 0\}$ and $\mathcal{N}(i) := \{c : \alpha_c(i) = 0\}$, with $K(i) := |\mathcal{P}(i)| \ll |\mathcal{C}|$.
That is, $\mathcal{P}(i)$ is the set of concepts visually present in image $i$, $\mathcal{N}(i)$ is the set of concepts absent from it, and $K(i)$ is the number of present concepts, which is typically much smaller than the total concept vocabulary $|\mathcal{C}|$.

Recall that for each concept $c$, we define the text coefficient $\beta_c(q) := f_{\mathcal{T}}(q)^\top e_c$, which measures how much the text embedding of query $q$ activates the direction of concept $c$.
Suppose the text query assigns uniform affirmed-concept margin $\gamma_+ > 0$ across present concepts and uniform negated-concept margin $\gamma_- > 0$ across absent concepts:
\begin{equation}
\beta_{c^+}({q}) \geq \gamma_+ \;\;\forall c \in \mathcal{P}(i), \qquad \beta_{c^{-}}({q}) \leq -\gamma_- \;\;\forall c \in \mathcal{N}(i).
\end{equation}
The unit-norm constraint $\sum_c \beta_c({q})^2 \leq 1$ then implies
\begin{equation}
K(i)\gamma_+^2 + (|\mathcal{C}| - K(i))\gamma_-^2 \leq 1,
\label{eq:budget-main}
\end{equation}
and consequently
\begin{equation}
|\gamma_-| \leq \sqrt{\frac{1 - K(i)\gamma_+^2}{|\mathcal{C}| - K(i)}} = \mathcal{O}\!\left(\frac{1}{\sqrt{|\mathcal{C}|}}\right).
\label{eq:budget-bound}
\end{equation}

Equation~\eqref{eq:budget-bound} complements Corollary~\ref{cor:beta-star-zero}: the conclusion $\beta^*_{c^-}(q) = 0$ arises both from optimization (the unit-norm maximizer of in-kind alignment) and from representation geometry (no unit vector can assign uniform negated concept margin across the dense set of absent concepts). Per-concept negated concept margin is compressed at rate $\mathcal{O}(1/\sqrt{|\mathcal{C}|})$ regardless of the training objective.

\subsection{Hard-negative supervision and frequency collapse}
\label{app:frequency}

We approximate the text encoder as token-additive: $f_{\mathcal{T}}(q) \approx \sum_{t \in \mathrm{tokens}(q)} W_t$. For negation queries $q = $ ``$c^+$ \emph{without} $c^-$'' sampled uniformly over $(c^+, c^-) \in \mathcal{C}^2$, the per-step participation rates are
\begin{itemize}
    \item ``without'' cue token: rate $1$ (appears in every negation example);
    \item concept token $c$: rate $1/|\mathcal{C}|$ in the negated position, dominantly trained by affirmation contexts elsewhere.
\end{itemize}
The learned embedding is therefore biased toward
\begin{equation}
f_{\mathcal{T}}(\text{``}c^+ \textit{ without } c^-\text{''}) \;\approx\; \underbrace{{\boldsymbol{e}}_{c^+}}_{\text{affirm}} + \underbrace{\boldsymbol{n}_{\mathrm{generic}}}_{\text{query-indep. offset}} + \underbrace{\mathcal{O}(1/|\mathcal{C}|)}_{\text{concept-specific}}.
\end{equation}

\paragraph{Generic offset is not concept-specific rejection.}
Although $\boldsymbol{n}_{\mathrm{generic}}$ may yield $\beta_{c^-}(q) < 0$, it fails to distinguish the negated concept $c^-$ from any unmentioned-but-allowed concept $r$. Faithful retrieval requires $\beta_{c^-}(q) < \beta_r(q)$ for all such $r$, but a query-independent offset penalizes $c^-$ and $r$ identically. 

This is empirically confirmed by NegBench's PCA visualization of ConCLIP, where `not(c)' embeddings collapse onte a single point regardless of $c$.~\citep{alhamoud2025vision}
Consequently, the conventional dual-encoder VLMs exhibit global conservatism, diluting the negated concepts rather than specifically rejecting them.

%
\section{Implementation details}
\label{app:impl}

\subsection{Hyperparameter settings}
\label{app:impl_hp}

\ourmethodshort\ uses one hyperparameters, all fixed across all backbones and task families. We do not tune any hyperparameter per backbone or per dataset.

\begin{table}[h]
\centering
\small
\caption{\ourmethodshort\ hyperparameters. All values fixed across backbones and tasks. $N$ denotes the candidate-pool size for the task ($N{=}4$ for MCQ, $N{=}|\mathcal{G}|$ for retrieval).}
\begin{tabular}{c c l}
\toprule
Symbol & Value & Description \\
\midrule
$\kappa$ & 50 & Sigmoid sharpness in $p_{c^-}(i)$ \\
$w_n$ & $1/\sqrt{N}$ & Rejection weight, scaled by candidate-pool size \\
\bottomrule
\end{tabular}
\label{tab:hp}
\end{table}

\subsection{Concept embedding}
\label{app:impl_concept}

Each concept $c$ in the vocabulary $\mathcal{C}$ is encoded as the L2-normalized average of four prompt templates:
\begin{equation}
\mathbf{e}_c \;=\; \textsc{normalize}\!\left( \frac{1}{4} \sum_{j=1}^{4} \mathrm{normalize}\!\bigl( f_{\mathcal{T}}(T_j(c)) \bigr) \right),
\label{eq:concept_emb}
\end{equation}
with templates $T_1(c){=}$``\emph{a photo of a \{$c$\}}'', $T_2(c){=}$``\emph{a picture of a \{$c$\}}'', $T_3(c){=}$``\emph{an image showing a \{$c$\}}'', and $T_4(c){=}$``\emph{a \{$c$\} in a photo}''.

\subsection{Negation parsing}
\label{app:impl_parse}

We use \texttt{spaCy}'s \texttt{en\_core\_web\_sm} for dependency parsing. A token is marked as a \emph{negation cue} if its dependency arc is \texttt{neg} or its lemma is in the cue set:
\begin{lstlisting}[basicstyle=\ttfamily\small,frame=single]
NEG_CUES = {"no", "not", "without", "nothing", "nowhere",
            "none", "never", "absent", "missing", "free of"}
\end{lstlisting}
The head noun of the dependency subtree rooted at the cue is taken as the negated concept $c^-$. Remaining noun phrases attached to the sentence root populate the affirmed concept set $\mathcal{C}^+$. The negation-stripped text $e_{\text{strip}}$ used in $\tilde{t}$'s multi-view averaging is constructed by removing all tokens in the negation subtree and re-joining with normalized whitespace and trailing punctuation.

\paragraph{Example.}
For the query \emph{``A car is present but no traffic light''}:
\begin{itemize}\itemsep0pt
    \item Negation cue: \texttt{no} $\Rightarrow$ subtree covers ``\emph{no traffic light}''
    \item $c^- = $ ``\emph{traffic light}''
    \item $\mathcal{C}^+ = \{\text{``\emph{car}''}\}$
    \item $e_{\text{strip}}$ encodes ``\emph{A car is present}''
\end{itemize}

\subsection{Image and video preprocessing}
\label{app:impl_imgproc}

We follow the standard CLIP preprocessing pipeline (resize, center-crop, normalize) at the backbone-specific input resolution (224 for B/32, 336 for L/14, 384 for SigLIP2 SO400M, etc.). For each image $x$, we encode both the original and its horizontal mirror, and use the L2-normalized mean of the two feature vectors:
\begin{equation}
f_{\mathcal{I}}(x) \;=\; \textsc{normalize}\!\left( \tfrac{1}{2} \bigl( f_{\mathcal{I}}^{\text{enc}}(x) + f_{\mathcal{I}}^{\text{enc}}(\textsc{flip}(x)) \bigr) \right).
\label{eq:img_flip}
\end{equation}
For MSR-VTT video retrieval, we extract four evenly-spaced frames per video (at quartile time positions) using \texttt{decord.VideoReader}. Each frame is encoded, and we use the L2-normalized mean of the four normalized frame features:
\begin{equation}
f_{\mathcal{I}}(v) \;=\; \textsc{normalize}\!\left( \frac{1}{4} \sum_{k=1}^{4} \textsc{normalize}~\!\bigl( f_{\mathcal{I}}^{\text{enc}}(v_k) \bigr) \right).
\label{eq:video_4frame}
\end{equation}

\section{Embedding correction reimplementation}
Aggarwal et al.~\cite{aggarwalseeing} report results only for CLIP ViT-B/32 and do not release code.
We re-implement their method following the procedure described in their paper: the negated concept direction is projected out from the query embedding with $\lambda = 1.9$, anchored to a neutral text embedding as described in their work.
Table~\ref{tab:multi-backbone-app} compares Embedding Correction and our method across all backbones.

\section{COCO-Confusion: Construction details}
\label{app:coco_conf}

This appendix details the construction of COCO-Confusion, our benchmark designed to evaluate whether a VLM can \emph{reject} a negated concept rather than merely dilute it. While Section~\ref{sec:experiments} introduces the benchmark, here we provide the full pipeline, statistics, and qualitative examples.

\subsection{Motivation: Why a new benchmark is neeeded}
\label{app:cc_motivation}

Existing negation retrieval benchmarks (e.g., NegBench) evaluate retrieval over a large unfiltered gallery (e.g., 5{,}000 COCO images). In such pools, the negated concept rarely co-occurs with the affirmed concept in the same image. As a result, even simple embedding correction-based methods perform competitively because there are few \emph{confusing} images to distinguish from correct ones. This makes existing benchmarks inadequate for measuring \emph{rejection}: without confusing distractors, dilution suffices.

COCO-Confusion is constructed precisely to expose this gap. For every query, the gallery is composed of images that all contain the affirmed concept, partitioned into:
\begin{itemize}\itemsep0pt
    \item \textbf{Correct images}: contain $c^+$ but \emph{not} $c^-$
    \item \textbf{Confusing images}: contain $c^+$ \emph{and} $c^-$
\end{itemize}
A method that merely dilutes $c^-$ scores both groups similarly. Only a method that genuinely \emph{rejects} $c^-$ separates them.

\subsection{Concept pair selection}
\label{app:cc_selection}

We use the 80 object categories of COCO from the \texttt{val2017} split, which contains 5{,}000 images with bounding-box annotations. For each image we extract its set of present categories from the official annotations.

\paragraph{Two-object pairs.}
For every ordered pair $(c^+, c^-)$ with $c^+ \neq c^-$, we partition images that contain $c^+$ into:
\begin{align*}
\mathcal{I}^+(c^+, c^-) &= \{\,i : c^+ \in i \,\wedge\, c^- \notin i\,\}, \\
\mathcal{I}^\pm(c^+, c^-) &= \{\,i : c^+ \in i \,\wedge\, c^- \in i\,\}.
\end{align*}
We retain a pair only if $|\mathcal{I}^+| \geq M_{\min}$ and $|\mathcal{I}^\pm| \geq M_{\min}$ with $M_{\min} = 10$. From each subset we sample up to $M_{\max} = 50$ images uniformly at random (seed~42). The query template is:
\begin{quote}
\emph{``A $c^+$ is present but no $c^-$.''}
\end{quote}

\paragraph{Three-object pairs.}
For every triple of distinct categories $(a, b, c)$ with at least $M_{\min}$ images containing all three, we generate three query partitions: $((a,b), c)$, $((a,c), b)$, and $((b,c), a)$, each treating two as affirmed and one as negated. Selection criteria are identical to the 2-object case applied to the partitioned subsets. The query template is:
\begin{quote}
\emph{``A $c^+_1$ and $c^+_2$ are present but no $c^-$.''}
\end{quote}

\paragraph{Coverage.}
The selection procedure covers 76 of 80 COCO categories in 2-object pairs and 62 of 80 in 3-object pairs (Table~\ref{tab:cc_coverage}). Categories absent from the benchmark are those with insufficient co-occurrence frequency to satisfy $M_{\min}{=}10$ in matched-pair construction (e.g., \emph{toaster}, \emph{hair drier}).

\begin{table}[h]
\centering
\small
\caption{Concept coverage in COCO-Confusion.}
\begin{tabular}{l c c}
\toprule
& 2-object & 3-object \\
\midrule
Total pairs (post-verification) & 950 & 1{,}740 \\
Unique affirmed concepts & 75 & 62 \\
Unique negated concepts & 76 & 62 \\
COCO categories used & 76/80 & 62/80 \\
\bottomrule
\end{tabular}
\label{tab:cc_coverage}
\end{table}

\subsection{OWLv2 verification}
\label{app:cc_owlv2}

COCO bounding-box annotations are accurate but may have boundary cases (heavy occlusion, partial visibility). To ensure clean labels, we re-verify every image--concept pair using OWLv2, an open-vocabulary object detector. We use the public checkpoint \texttt{google/owlv2-base-patch16-ensemble} with text queries of the form ``\emph{a photo of a $c$}.''

For each image $i$ and concept set $\mathcal{C}_q = \{c^+\} \cup \{c^-\}$, OWLv2 produces a maximum detection confidence $\text{score}_c(i) \in [0,1]$ per concept. We retain images according to the following rules at threshold $\tau_{\text{OWL}} = 0.1$:
\begin{itemize}\itemsep0pt
\item \textbf{Correct image kept} iff $\max_{c \in \mathcal{C}^+} \text{score}_c(i) \geq \tau_{\text{OWL}}$ \emph{and} $\text{score}_{c^-}(i) < \tau_{\text{OWL}}$.
\item \textbf{Confusing image kept} iff $\max_{c \in \mathcal{C}^+} \text{score}_c(i) \geq \tau_{\text{OWL}}$ \emph{and} $\text{score}_{c^-}(i) \geq \tau_{\text{OWL}}$.
\item \textbf{Pair kept} iff $\geq 5$ verified images remain in both subsets.
\end{itemize}
The low threshold $\tau_{\text{OWL}}{=}0.1$ favors recall, reducing the risk of falsely classifying a partially-visible $c^-$ object as ``absent.''

\paragraph{Retention rates.}
Pair-level retention is high: 99.9\% (2-object) and 97.1\% (3-object), consistent with COCO annotations being already accurate. Filtering primarily refines the per-pair image lists rather than discarding pairs.

\begin{table}[h]
\centering
\small
\caption{COCO-Confusion final statistics.}
\begin{tabular}{l c c}
\toprule
& 2-object & 3-object \\
\midrule
Total pairs & 950 & 1{,}740 \\
\bottomrule
\end{tabular}
\label{tab:cc_stats}
\end{table}

\subsection{Evaluation protocol}
\label{app:cc_eval}

For each query, the retrieval gallery is the union of its verified correct and confusing images. We compute mean Average Precision (mAP) over the gallery, with correct images as ground-truth positives. The benchmark-level metric is the macro-average of per-pair mAP across the 950 (Conf-2C) or 1{,}740 (Conf-3C) pairs.

\subsection{Qualitative examples}
\label{app:cc_examples}

\begin{figure}[t]
\centering
\setlength{\fboxsep}{0pt}%
\setlength{\fboxrule}{1pt}%

\begin{subfigure}{0.175\textwidth}\centering
  \fcolorbox{correctgreen}{white}{\includegraphics[width=\linewidth]{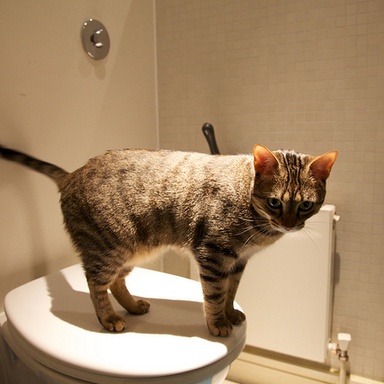}}
\end{subfigure}\hfill
\begin{subfigure}{0.175\textwidth}\centering
  \fcolorbox{correctgreen}{white}{\includegraphics[width=\linewidth]{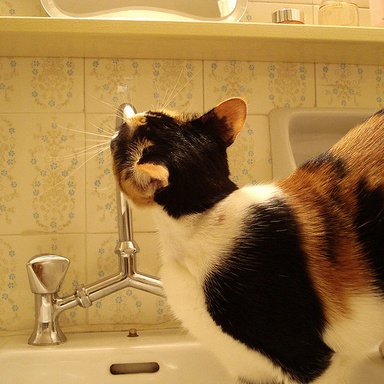}}
\end{subfigure}\hfill
\begin{subfigure}{0.175\textwidth}\centering
  \fcolorbox{correctgreen}{white}{\includegraphics[width=\linewidth]{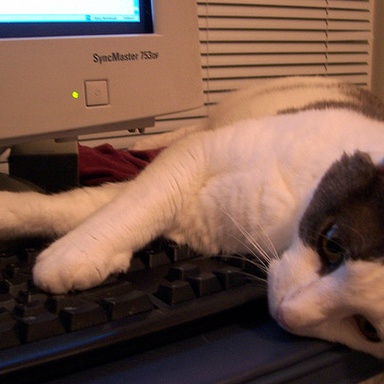}}
\end{subfigure}\hfill
\begin{subfigure}{0.175\textwidth}\centering
  \fcolorbox{correctgreen}{white}{\includegraphics[width=\linewidth]{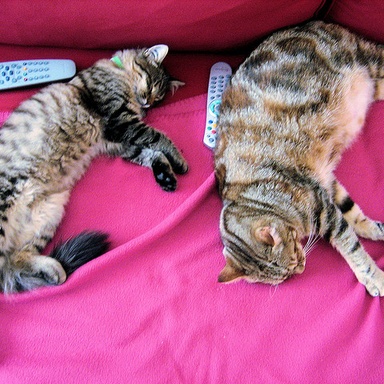}}
\end{subfigure}\hfill
\begin{subfigure}{0.175\textwidth}\centering
  \fcolorbox{correctgreen}{white}{\includegraphics[width=\linewidth]{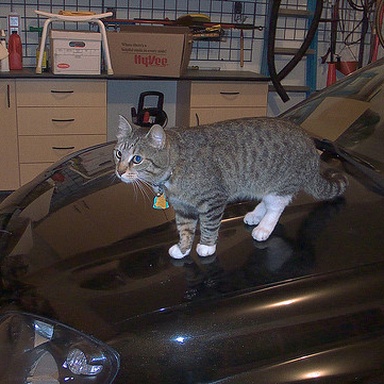}}
\end{subfigure}

\vspace{2pt}
{\small \textcolor{correctgreen}{\textbf{Correct}: cat present, no dog}}
\vspace{6pt}

\begin{subfigure}{0.175\textwidth}\centering
  \fcolorbox{confusingred}{white}{\includegraphics[width=\linewidth]{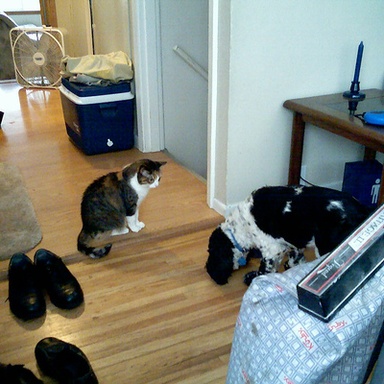}}
\end{subfigure}\hfill
\begin{subfigure}{0.175\textwidth}\centering
  \fcolorbox{confusingred}{white}{\includegraphics[width=\linewidth]{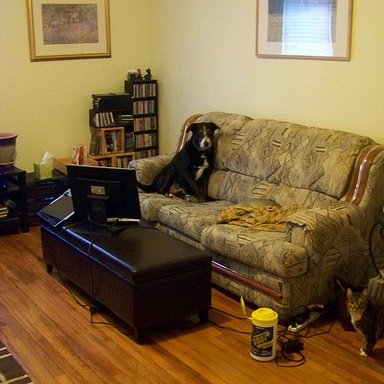}}
\end{subfigure}\hfill
\begin{subfigure}{0.175\textwidth}\centering
  \fcolorbox{confusingred}{white}{\includegraphics[width=\linewidth]{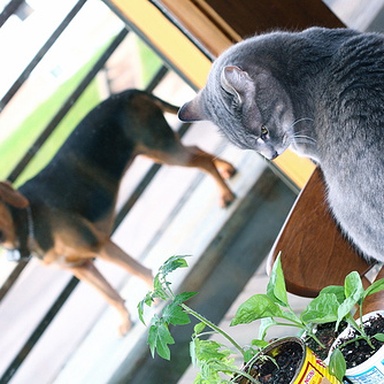}}
\end{subfigure}\hfill
\begin{subfigure}{0.175\textwidth}\centering
  \fcolorbox{confusingred}{white}{\includegraphics[width=\linewidth]{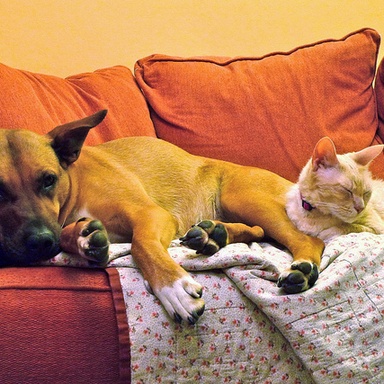}}
\end{subfigure}\hfill
\begin{subfigure}{0.175\textwidth}\centering
  \fcolorbox{confusingred}{white}{\includegraphics[width=\linewidth]{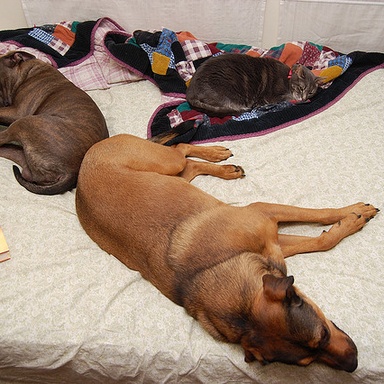}}
\end{subfigure}

\vspace{2pt}
{\small \textcolor{confusingred}{\textbf{Confusing}: cat present, dog also present}}

\caption{2-object COCO-Confusion example. \emph{Query: ``a cat is present but no dog.''} Top row: correct images, containing a cat but no dog. Bottom row: confusing images, containing both a cat and a dog. All ten images contain a cat---only the negated concept differs across rows. A method that merely dilutes the ``dog'' direction in the text embedding scores all ten images similarly. Only a method that genuinely rejects $c^-$ separates correct from confusing.}
\label{fig:cc_examples_2c}
\end{figure}

\begin{figure}[t]
\centering
\setlength{\fboxsep}{0pt}%
\setlength{\fboxrule}{1pt}%

\begin{subfigure}{0.175\textwidth}\centering
  \fcolorbox{correctgreen}{white}{\includegraphics[width=\linewidth]{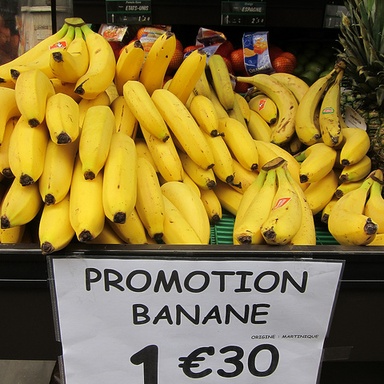}}
\end{subfigure}\hfill
\begin{subfigure}{0.175\textwidth}\centering
  \fcolorbox{correctgreen}{white}{\includegraphics[width=\linewidth]{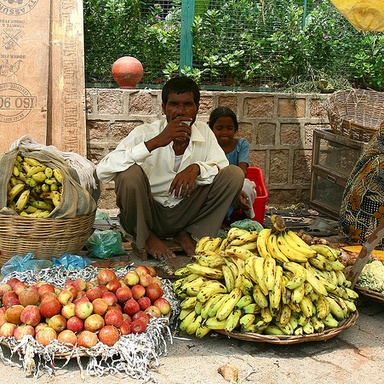}}
\end{subfigure}\hfill
\begin{subfigure}{0.175\textwidth}\centering
  \fcolorbox{correctgreen}{white}{\includegraphics[width=\linewidth]{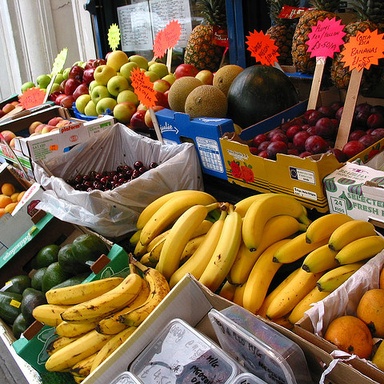}}
\end{subfigure}\hfill
\begin{subfigure}{0.175\textwidth}\centering
  \fcolorbox{correctgreen}{white}{\includegraphics[width=\linewidth]{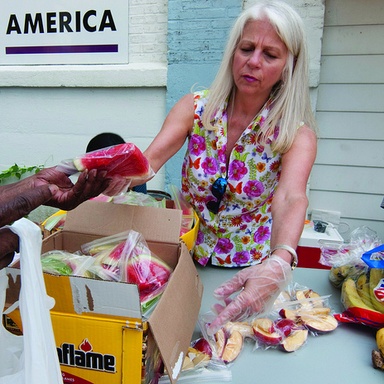}}
\end{subfigure}\hfill
\begin{subfigure}{0.175\textwidth}\centering
  \fcolorbox{correctgreen}{white}{\includegraphics[width=\linewidth]{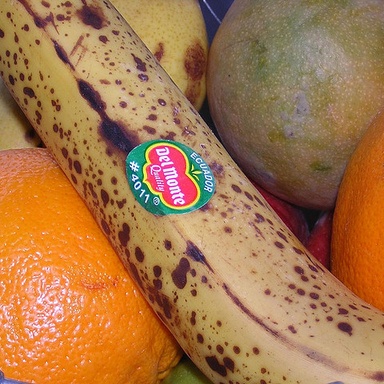}}
\end{subfigure}

\vspace{2pt}
{\small \textcolor{correctgreen}{\textbf{Correct}: apple\,+\,banana present, no bowl}}
\vspace{6pt}

\begin{subfigure}{0.175\textwidth}\centering
  \fcolorbox{confusingred}{white}{\includegraphics[width=\linewidth]{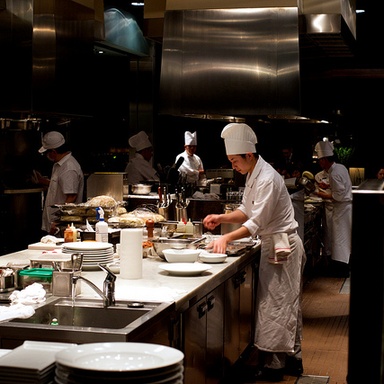}}
\end{subfigure}\hfill
\begin{subfigure}{0.175\textwidth}\centering
  \fcolorbox{confusingred}{white}{\includegraphics[width=\linewidth]{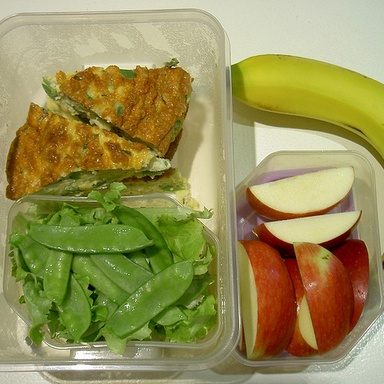}}
\end{subfigure}\hfill
\begin{subfigure}{0.175\textwidth}\centering
  \fcolorbox{confusingred}{white}{\includegraphics[width=\linewidth]{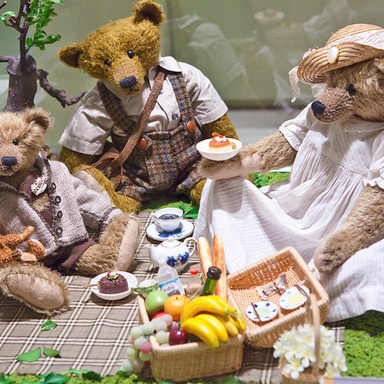}}
\end{subfigure}\hfill
\begin{subfigure}{0.175\textwidth}\centering
  \fcolorbox{confusingred}{white}{\includegraphics[width=\linewidth]{}}
\end{subfigure}\hfill
\begin{subfigure}{0.175\textwidth}\centering
  \fcolorbox{confusingred}{white}{\includegraphics[width=\linewidth]{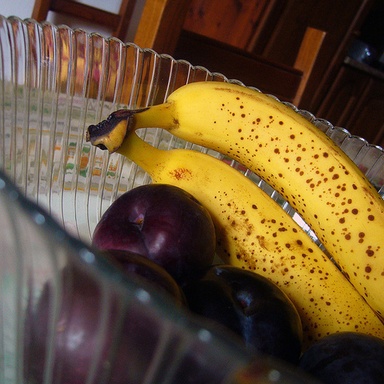}}
\end{subfigure}

\vspace{2pt}
{\small \textcolor{confusingred}{\textbf{Confusing}: apple\,+\,banana present, bowl also present}}

\caption{3-object COCO-Confusion example. \emph{Query: ``an apple and banana are present but no bowl.''} The query requires affirming both apple and banana while rejecting bowl. All shown images contain an apple and a banana. Only the bottom row also contains a bowl.}
\label{fig:cc_examples_3c}
\end{figure}

Figure~\ref{fig:cc_examples_2c} shows a 2-object pair (\emph{cat / dog}). Figure~\ref{fig:cc_examples_3c} shows a 3-object pair (\emph{apple\,+\,banana / bowl}). In each figure, the top row shows verified correct images (green border, contain the affirmed concept(s) but not the negated concept) and the bottom row shows verified confusing images (red border, contain both the affirmed and negated concepts).

\subsection{What makes COCO-Confusion different}
\label{app:cc_unique}

Compared to NegBench standard retrieval (1 query against a 5{,}000-image gallery dominated by irrelevant images):
\begin{enumerate}\itemsep0pt
    \item \textbf{Curated galleries with high confusing density.} Each pair gallery contains 50--100 images, of which 20--50\% contain the negated concept. Standard retrieval has $<1\%$ confusing images per query.
    \item \textbf{OWLv2 cross-validation.} Every image label is double-verified by an external open-vocabulary detector, eliminating bounding-box edge cases.
\end{enumerate}
Together these properties make COCO-Confusion a stress test for \emph{rejection}, while standard retrieval benchmarks remain useful for measuring \emph{dilution-level} negation handling.

\end{document}